\documentclass[twocolumn, 11pt]{ieeeconf}

\usepackage{amsmath,amssymb,bm}

\usepackage{xcolor}
\usepackage{graphicx}
\graphicspath{{figures/}}
\usepackage{accents}

\newcommand{\sConf}{\mathcal{Q}}
\newcommand{\aG}{\mathsf{g}}

\newcommand{\gop}{\circ}
\newcommand{\groupderiv}[2][]{\accentset{\scriptstyle\gop}{#2}}

\newcommand{\fiberspace}{G}					
\newcommand{\basespace}{R}
\newcommand{\concept}[1]{\emph{#1}}

\usepackage{authblk}

\usepackage[sorting=none]{biblatex}
\begin{document}
\title{Optimizing Gait Libraries via a Coverage Metric}

\author[1,2]{Brian Bittner}
\author[1,3]{Shai Revzen}
\affil[1]{Robotics Institute, University of Michigan, Ann Arbor, USA}
\affil[2]{Johns Hopkins University Applied Physics Lab, Laurel, MD, USA \\ brian.bittner@jhuapl.edu}
\affil[3]{Electrical Engineering and Computer Science Department  \& Ecology and Evolutionary Biology Department, University of Michigan, Ann Arbor, USA \\ shrevzen@umich.edu}

\maketitle

\begin{abstract}
Many robots move through the world by composing locomotion primitives like steps and turns.
To do so well, robots need not have primitives that make intuitive sense to humans.
This becomes of paramount importance when robots are damaged and no longer move as designed.
Here we propose a goal function we call ``coverage'', that represents the usefulness of a library of locomotion primitives in a manner agnostic to the particulars of the primitives themselves.
We demonstrate the ability to optimize coverage on both simulated and physical robots, and show that coverage can be rapidly recovered after injury.
This suggests that by optimizing for coverage, robots can sustain their ability to navigate through the world even in the face of significant mechanical failures.
The benefits of this approach are enhanced by sample-efficient, data-driven approaches to system identification that can rapidly inform the optimization of primitives.
We found that the number of degrees of freedom improved the rate of recovery of our simulated robots, a rare result in the fields of gait optimization and reinforcement learning.
We showed that a robot with limbs made of tree branches (for which no CAD model or first principles model was available) is able to quickly find an effective high-coverage library of motion primitives.
The optimized primitives are entirely non-obvious to a human observer, and thus are unlikely to be attainable through manual tuning.
\end{abstract}

\section{Introduction}
One of the most common sub-problems in modern robotics is path-planning, and the choice of path is usually framed as a precise or approximate optimal control problem.
When restricted to mobile robots moving through many practical environments, the path planning problem enjoys an additional important symmetry.
Given the configuration of the robot body, the short-horizon movements it can execute are the same at nearly every point in space.
This allows short time horizon \concept{primitives} to be optimized offline and pre-cached, later to be composed sequentially to produce solutions to the full path planning problem.
For example, a humanoid robot such as ATLAS can execute the same walking steps at any point on flat, unobstructed ground.
To plan the motions of the robot walking through a building, one can sequence primitives for generating a collection of steps in the correct order instead of solving the full high-dimensional planning problem.

Unfortunately, the primitives seen in such library-based plans are usually created by hand, and generated with constraints that help reduce the complexity of an individual planning problem.
For example, a common choice for 2D motion, dating back to the turtle robots of the 1950s \cite{walter1953living}, is to have linear translation and turning in place as primitives.
However, this particular choice for generating movements is entirely arbitrary.
A given robot may be far more efficient moving diagonally or turning while moving on an arc.
The ability to optimize for a library of useful primitives can come to have critical importance when a robot is damaged, and the choice of best available primitives might no longer correspond to any motion obvious to a human operator.

Here we present a method to optimize an entire primitive library concurrently so as to achieve the ability to efficiently plan over the space of body motions with that library.
By optimizing for the \concept{coverage} goal function we define, the library selected will be able to express desired short-horizon plans through composition of primitives from the library.

One approach for approximate optimal planning is to construct a state lattice \cite{pivtoraiko2009differentially, knepper2009path} -- a discrete collection of states that can be generated by a library of primitives.
Planning consists of sequencing primitives to travel along the lattice to approximate the total desired motion.
Such previous work on state lattices suggests that a good collection of primitives are:
\begin{itemize}
\item \textit{complete} -- the space of desirable motions is densely populated
\item \textit{fast-to-compute} -- the robot is able to select and use primitives in real time
\item \textit{path optimal} -- each individual primitive should be similar to a globally optimal path available between its start and end states.
\end{itemize}
When generating primitives, one has a variety of options to chose from \cite{fod2002automated, frazzoli2005maneuver, schaal2005learning, hauser2008using, pivtoraiko2011kinodynamic}.
Strategies can include learning from demonstration as well as prioritizing spatial properties of the output trajectories of the system.
Large primitive libraries are often winnowed down to save run time or increase the planning update rate.
Our work can be viewed (in part) as a means for generating very small, very expressive libraries of primitives.

Our work can also be seen as a way to relax the standard assumption used in optimizing gaits, namely pre-specifying the direction \cite{hatton2010optimizing, gong2016simplifying} or turn rate of motion \cite{da20162d, hartley2017stabilization} over a single cycle.
We observe that most of the value a primitive has is not intrinsic, but rather in its contribution to support other compositions of primitives available to a planner and the overall needs of the planning task.
We thus provide a way to evaluate libraries of primitives rather than their individual characteristics.
Primitives that have negligible exploration value in isolation may be critical to more densely maneuvering through space.
We demonstrate how our coverage measure values such primitives rather than discards them.

Using our approach is nearly paradigmatically opposite to traditional behavior learning in robotics.
We allow for the optimizer to ``ask'' the robot what ways are convenient to move, rather than dictating how the robot should move \textit{a-priori}.
A subsequent advantage is that mechanical designers can rethink common design criteria for locomotors.
Typically robots acting on a planar workspace are designed to have at least one mode by which gaits translate the system without rotating it.
This preference may simply be the result of an anthropocentric bias.
It is how humans move to avoid disorientation and dizziness, but it is not a universal requirement for effective locomotion.
The coverage measure, being devoid of such biases, allows a broader range of robot mechanisms to score highly.
Crucially, it can also potentially allow broken robots to recover their ability to plan motions by rapidly regenerating a primitive library while damaged.
%
\subsection{Overview of this paper}
Below we briefly review Lie groups in \S\ref{sec:groups}, so as to use them to represent the composition of primitive libraries as a sequence of group actions acting on a Lie group of body locations.
Using this representation, we define coverage in \S\ref{sec:coverage} and provide examples of how it can be computed on the rigid body groups $\mathsf{SE}(2)$ and $\mathsf{SE}(3)$.
In \S\ref{sec:toy-examples} we use this coverage to discuss a collection of toy systems whose locomotion ability becomes easy to appreciate through our approach.
We translate this framework of primitive optimization to the world of gait driven systems in \S\ref{sec:geomech}.
There we pay special attention to highly damped systems, where the task of chaining primitives can be greatly simplified.
We present coverage optimization of gait libraries for some Purcell swimmer models in \S\ref{sec:covopt}.
Using coverage as a tool, in \S\ref{sec:recovery} we investigate the ability of the Purcell swimmer to recover from joint locking failures.
Finally, we emphasize the ability of the optimization to work on unintuitive robots, even when we do not specify the robot kinematics, mass distribution, or material properties.
We demonstrate with a robot whose limbs are made of tree branches, and which gained the ability to navigate on the floor with less than eleven minutes of hardware-in-the-loop optimization for coverage.

\section{Expressing Motion Through the Space of Discrete Actions}
\label{sec:groups}

\begin{figure}
\begin{center}
\includegraphics[width=.5\textwidth]{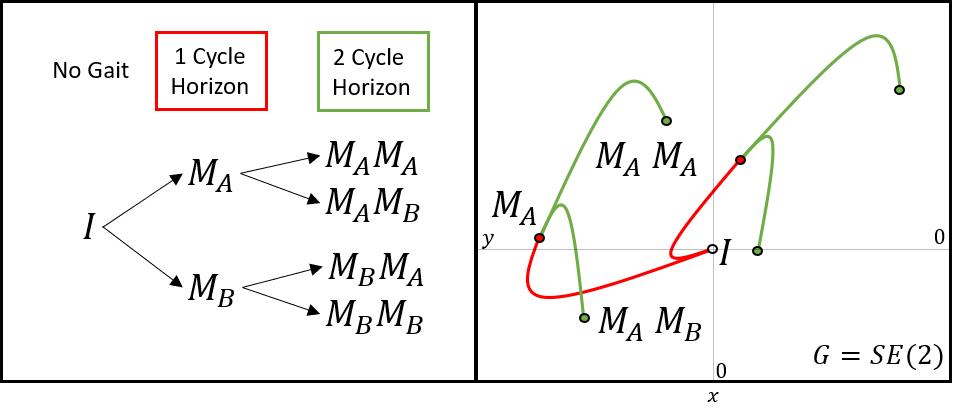}
\caption{ %
  Illustration of composing gait cycles.
  Here, the two group actions ($M_A, M_B$) are applied in various orders and combinations. %
  An $n$-step finite horizon planner considers words, a concatenation of group action letters, of length $n$. %
  For a two letter action library, $n$ step planners consider $2^n$ paths (trees in the left panel). %
  We illustrated a possible case of such motions. %
  By assuming that the robot is oriented tangent to the direction of motion, the resulting motions can be represented by their projection on the translational plane (right panel).}
\label{fig:words}
\end{center}
\end{figure}

To represent motion, we assumed that the configuration space $\sConf$ of our moving robot could be factored as a product of a shape $\basespace$ and a (generalized) position $\fiberspace$.
This generalized position is a Lie group, typically a sub-group of the rigid body motions $\mathsf{SE}(3)$.
In this work, we restricted our attention to motion in the ground plane, $\mathsf{SE}(2)$.
We produced motion using \concept{(periodic) gaits}, which we take to mean periodic changes in shape that produce a predictable body motion.
Formally, a gait $b$ is a function $\gamma_b: S^{1} \to \basespace$ that produces a body motion $M_b\in \fiberspace$.
A stride consisting of a left step followed by a right step is an example of a gait cycle of human walking.
Given a finite selection of gaits and a means for switching between them, a planner can produce any motion that corresponds to a word composed of the group elements (letters) those gaits generate.
For example, with gaits $\gamma_a$ and $\gamma_b$, and provided any sequence is allowed, one can produce the motions ($I, M_a, M_b, M_a^2, M_a M_b, M_b^2, M_b M_a, M_a^3, \ldots$).
Figure \ref{fig:words} provides a visualization of what this representation looks like for motion planning in a planar workspace.

In this paper, we restricted our discussion to primitive libraries consisting of single cycles of different gaits as the primitives and assumed the gaits are connected in internal state at their start and end configuration.
This is not generally the case, and receives more careful treatment in \S\ref{sec:geomech}.

\section{Specifying the Loss Function}
\label{sec:coverage}
As conventionally practiced, a motion planner is given some parameters $x$, a means to generate motions $M(x) \in \fiberspace$, and some loss function which it will minimize.
Because we did not consider power efficiency here, we took the loss function
\begin{equation}
\tilde \eta: \fiberspace \to \mathbb{R}^+
\end{equation}
to be purely a function of the endpoint.
Including additional factors in the loss functions for individual primitives is only a matter of book-keeping, provided the loss function of the overall path is additive in those of its constituent primitives.
We assumed that the loss function is written relative to some desired goal position $G$ of the motion, and defined a relative (local) loss function using the Lie algebra
$
\eta(\xi) := \tilde \eta( \exp(\xi) G ).
$ 
We then optimized for the parameters $x$ of the primitive with respect to the loss function
$
x \mapsto \eta \circ \log( M(x) G^{-1} ).
$ 
Any left invariant distance metric for $\mathsf{SE}(2)$ or $\mathsf{SE}(3)$  provides practical implementation of $\tilde \eta$.
Picking such a metric boils down to a choice of a constant that relates the loss of translation errors to the loss of orientation errors, and thus this choice is application specific.
In this work, we chose this parameter to make a half rotation on any axis to be of equal loss to displacement of a body length.

We set up our optimization as follows: let
$
\mathbf{G} := \{ G_i \}_{i=1}^n
$ 
be a set of goal motions and
$
\mathbf{W} := \{ w_i \}_{i=1}^n \subset \mathbb{R}^+
$ 
be a corresponding set of weights.
Let
$
\mathbf{M} := \{ M_j \}_{j=1}^m
$ 
be a set of achievable motions.

We defined the coverage cost
\begin{equation}
h(\mathbf{M}) := \sum_i w_i \min_j \eta \circ \log (M_j G^{-1}_i).
\end{equation}
We further defined $h_k(\mathbf{M})$ as the cost of the set of words of $k$ or fewer elements of $\mathbf{M}$.
The coverage cost is the sum of the costs of the best approximations available for $G_i$, given the achievable $M_j$ and weighted by the weights $w_i$ for each $G_i$.

\begin{figure}
\begin{center}
\includegraphics[width=.5\textwidth]{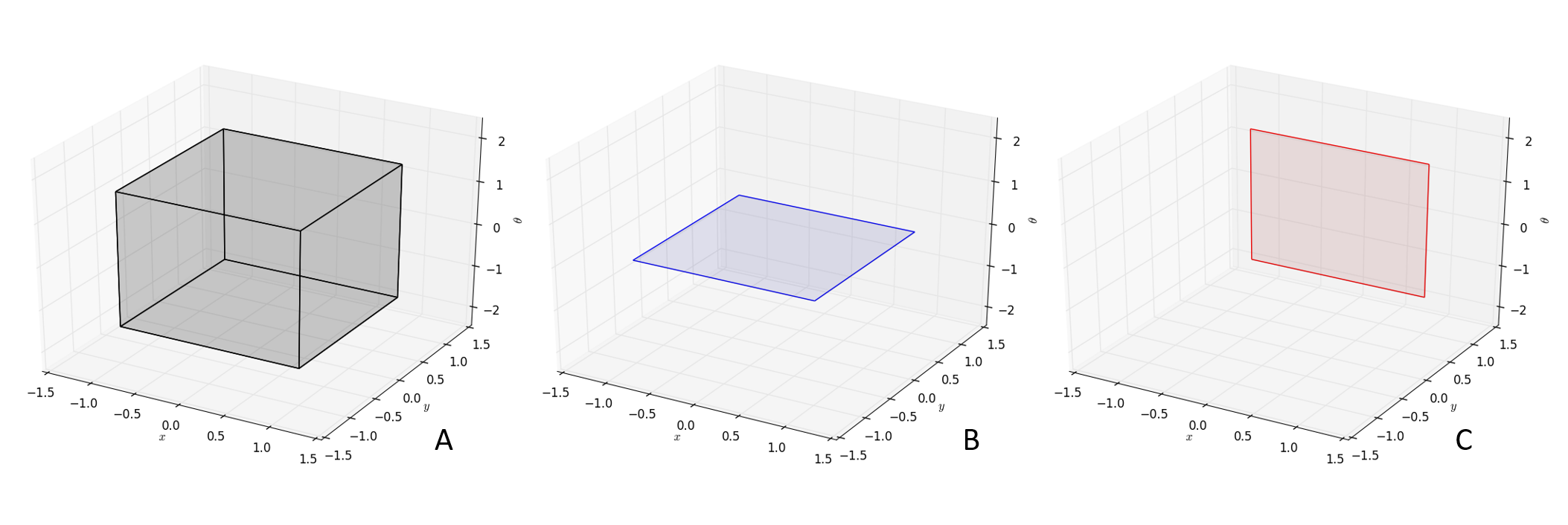}
\caption{ Expressive power of the coverage cost. %
  One has a variety of choices for placement and weighting of coverage points. %
  We provided some suggestions for various design goals on the space of planar rigid body motions. %
  A user can prioritize versatility (panel A), zero-rotation translation (panel B), or right lateral movement (panel C). %
  Volumes and planes are suggested regions for the user to evenly distribute uniformly weighted coverage points $G_i$. %
} \label{fig:cov-vis}
\end{center}
\end{figure}

\subsection{Higher order maneuvers}
\label{sec:commutator}

One of the surprising insights of nonlinear control is that the non-commutativity of control actions can make reachable the iterated Lie brackets of a control distribution \cite{sastry1993structure,sastry2013nonlinear}.
The discrete primitive library equivalent of this insight is the observation that the \concept{commutator} word $M_a M_b M^{-1}_a M^{-1}_b$ can at times reach directions that no word of the form $M_a^n M_b^m$ could reach.
Thus, designing $h_k(\cdot)$ such that $k \geq 4$ allows these higher-order maneuvers to be included.
It is, however, important to note that the coverage computation time scales exponentially with $k$.
For this reason, we used $k=4$ in our implementations here.

\subsection{Design choices for coverage points}

The coverage cost presented offers a user the ability to specify both the placement and weighting of coverage points.
The selection of the points and weights can radically change the priorities of the optimizer.
A user prioritizing versatility may want the robot to be able to reach all parts of its local position space.
They can place a uniformly weighted set of points distributed evenly within some volume around the identity (non-)motion (Figure \ref{fig:cov-vis} A).
Another user may wish to find a combination of gaits that translate while preserving orientation.
That might correspond to a coverage point distribution in a thin wedge near the 2D slice of $\mathsf{SE}(3)$ corresponding to no rotation (Figure \ref{fig:cov-vis} B).
Such maneuvers might be useful for an inspection robot that needs to maintain a visual field of view while moving.
If one had a more specific navigational goal, \textit{e.g.} finding a way to translate laterally to the right, such a goal can also be captured (Figure \ref{fig:cov-vis} C).

The $w_i$ weighted collection of coverage goal points $G_i$ can be seen as a discrete approximation to a measure on the group.
Increasing the number of goal points in a region while keeping the total weight constant implies a preference for higher resolution in that region.
Changing the weight while keeping the goal points unchanged implies an increase or decrease in the importance of approximating those goal motions with the primitive library.

\begin{figure}
\begin{center}
\includegraphics[width=.5\textwidth]{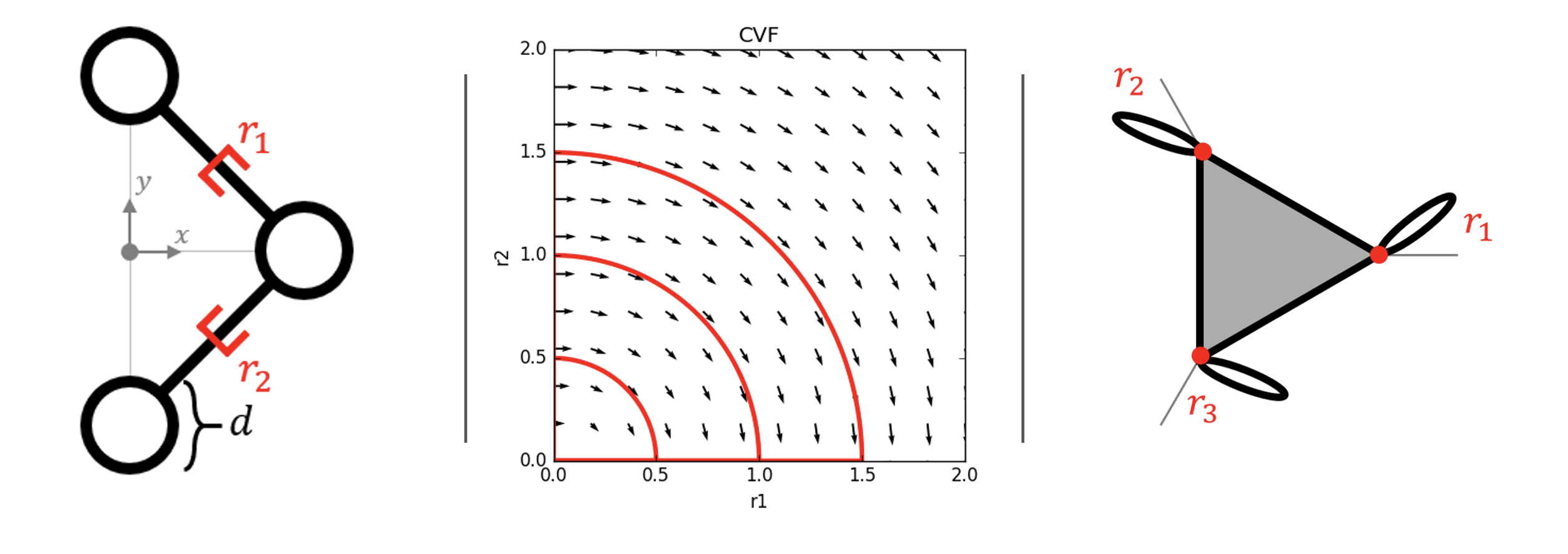}
\caption{Here we describe two mechanical systems that may appear as unconventional travelers.
The two-slider swimmer (left) can move spheres along prismatic joints.
The motion simultaneously induces a thrust on the system while changing the geometry of drag forces acting on the system.
We plotted the gaits selected for the two-slider swimmer on the rotational connection vector field \cite{hatton2010optimizing} of the two-slider swimmer (middle).
This provided insight into how shape change can influence body velocity.
We can see that paths (shown in red) that start in the corner at the origin, travel along a shape axis, sweep at a constant radius to another axis, then return to the origin.
The connection vector field aided gait selection of the two-slider swimmer, which is discussed in \S\ref{sec:select-twoslide}.
The three-branch swimmer (right) has two-joints that can rotate, fixed to the end of a triangle.
Since the shape space of the three-branch swimmer is not restricted to planar representations, we selected gaits in a different way.
}
\label{fig:toysys}
\end{center}
\end{figure}

\section{Coverage Invites Non-traditional Mechanical Designs}
\label{sec:toy-examples}

\begin{figure*}
\begin{center}
\includegraphics[width=\textwidth]{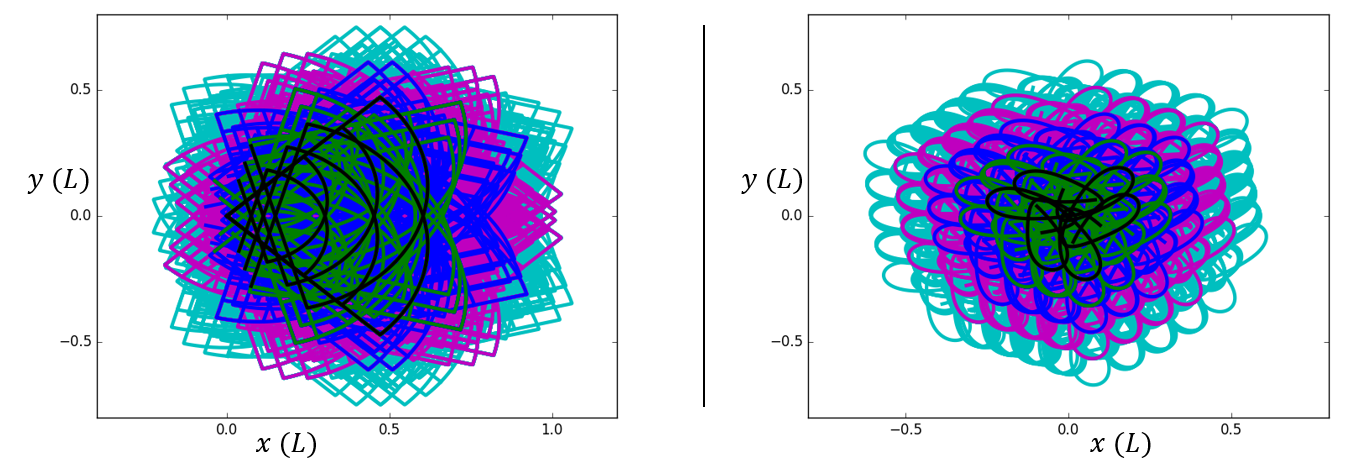}
\caption{Both systems were able to explore their local environments in a way that is unrestricted to translation in the plane.
We plotted paths to show the number of steps required to arrive at a target pose, omitting visualization of the orientation ($\theta$) component of the full $\mathsf{SE}(2)$ pose.
We plotted motions available in 5 steps (1=black, 2=green, 3=blue, 4=magenta, and 5=cyan).
At 5 steps (cyan), the system had a broad variety of poses at its disposal.
Both systems appear to be capable of navigating through environments with sparse obstacles.}
\label{fig:swimhoriz}
\end{center}
\end{figure*}

Here we intentionally designed two robots that move in unconventional ways.
The first cannot translate without rotating.
The second has a trilateral symmetry.
Often, roboticists do not consider such systems because their mobility is non-intuitive.
Yet we have shown below that both systems can move quite effectively in $\mathsf{SE}(2)$.

\subsection{Introducing two new mechanisms}
Both of the mechanical system models we present are swimmers that operate at the limit of low Reynolds number fluid dynamics \cite{purcell1977life}, where friction dominates inertia.
The motion of these systems is fully dominated by the drag forces induced by the internal velocities of the robots shape variables $r \in R$ and body velocities $ \groupderiv{g} \in \mathsf{T} \mathsf{SE}(2)$.
The motions of these systems can be usefully inspected using the tools of \cite{hatton2010optimizing, hatton2013geometric, ramasamy2016soap}.

\subsubsection{Two-slider swimmer model:}
The two-slider swimmer in Figure \ref{fig:toysys} moves via the prismatic joints driven by strictly positive displacements $r_1$ and $r_2$.
The viscous force on each sphere is linear in translational velocity and cubic in rotational velocity.
Its full model is:
\begin{align}
    &\begin{bmatrix} 3d & 0 & 0 \\ 0 & 3d & 0 \\ -d r_2 & -d r_1 & d ({r_1}^2 + {r_2}^2) + \frac{d^3}{4} \end{bmatrix} \groupderiv{g} = R(\alpha) \begin{bmatrix} 0 & d \\ -d & 0 \\ 0 & 0 \end{bmatrix} \dot{r} \\
    &\alpha = -\frac{\pi}{2} \nonumber
\end{align}
where R takes input parameters to a rotation about the origin on $\mathsf{SE}(2)$.

\subsubsection{Three-branch swimmer model:}
We also designed the three-branch swimmer (see Figure \ref{fig:toysys}), another viscous swimmer.
two-joints are free to rotate from the points of the triangle.
For biological intuition for how a system like this might move, a starfish might move like a pentagonal five-branch system with longer segments of links at each vertex.
The links interact via the slender body theory of Cox \cite{cox1970motion}, the same that was used for the swimmer in \cite{hatton2013geometric} and paddles in \cite{kvalheim2019gait}.
The drag of the triangular piece is represented by three static links that point from the center of the triangle to their respective attachment points.

\subsection{Hand selecting gaits}

\subsubsection{Gait selection for two-slider swimmer:} \label{sec:select-twoslide}

By inspection of the connection vector field of the rotational component of the two-slider swimmer (see Figure \ref{fig:toysys}), we saw that a variety of turning modes could be excited.
Hand-selected gaits all started at the origin of the base space, travel along the axis of one shape variable, then translated at a constant radius from the origin, traveling from one positive end of a shape axis to the other.
Each path was then sent to the origin via the other shape variable.
We can see from the curl of the vector field that clockwise gaits will yield positive rotation, and counter-clockwise gaits will yield negative rotation.
The three paths printed in red represent three magnitudes of turning the system can choose.
The larger the radius, the greater the turn will be, as explained in Figure \ref{fig:toysys}.
Each gait also induces a translational displacement of the system from its starting location.

\subsubsection{Gait selection for three-branch swimmer:}

The three-branch swimmer is less amenable to inspection by the connection vector field methods since it has a third shape variable.
Reduction methods (such as \cite{rieser2019geometric}) can make such gait analysis useful for more complex Stokesian systems.
Geometric gait optimization can also be employed on this analytical system to obtain a collection of gaits, maximizing various objective functions \cite{ramasamy2019geometry}.
In this example, our goal is to demonstrate that sub-optimal behaviors can be interpreted as valuable through the coverage metric.
Therefor, we selected gaits rather heuristically.
Two criteria for selected motions were to avoid self-intersections and enclose a non-zero volume in the shape space\footnote{The scallop theorem \cite{lauga2011life} ensures that gaits with zero enclosed volume will achieve zero displacement in the stokes regime}.
Two links oscillated in anti-phase, providing a thrust that acts through a line from the midsection of their attachment points to the third link's attachment point.
The third link oscillated out of phase by a quarter cycle.
We designed the gaits as:

\begin{align}
r_{mod(k,3)+1} &= \sin(\varphi) \\
r_{mod(k+1,3)+1} &= 1-\cos(\varphi) \\
r_{mod(k+2,3)+1} &= -1+\cos(\varphi)
\end{align}
for $\varphi \in S^1$ with gaits $\gamma_k$ enumerated k = (1,2,3).
These three gaits generate three group actions, which can also be run backward in $\varphi$, generating three inverse group actions.

Each system had six gaits at its disposal.
By inspection of Figure \ref{fig:swimhoriz}, we observe the local planning ability of the systems, only using the six gaits as possible actions (letters) of their total motion (word).
We highlight the key takeaway of this section.
Behaviors that were not useful in isolation were critical to providing dense coverage.
Furthermore, these behaviors may lie outside the scope of typical behaviors that a roboticist may prescribe for a system.

\section{Connecting Gait and Motion}
\label{sec:geomech}

The algebraic structure for computing available motions is straightforward: separate gaits were concatenated as a string of group multiplications.
What dynamical properties were required for such assumptions?
We cover the assumptions we made in this section, using the language of geometric mechanics.
For general dynamical systems, combining gaits would require a transition behavior that matches the internal state ($r,\dot{r},p$) of the endpoint of one gait and connects it with the internal state ($r,\dot{r},p$) of the starting point of the next gait.
There exist a class of systems where the matching requirements are highly relaxed.

\subsection{Planning simplifications in principally kinematic systems}
\label{sec:stokes}

The class of systems we focused on in this work inhabit the Stokes regime \cite{kelly1995geometric}, which encompass the dynamic qualities of the principally kinematic case covered in \cite{ostrowski1998geometric}.
A well known example of such systems is low Reynolds number swimmers \cite{kelly1996geometry, shapere1989geom}.
However, we recently accumulated evidence that this theory applies to multi-legged locomotion \cite{wu2019coulomb, clifton2020uneven}.
The function $A(\cdot)$ connects gaits, as body shape loops, to the motion they induce, called the ``holonomy'' of the loop.

It is known from Stokes' Theorem that a closed loop integral of a vector field is equal to the area integral of the volume enclosed by the loop.
This is an approximation of the motion for non-abelian systems, and careful selection of coordinates can improve the quality of this approximation \cite{hatton2011geometric}.
This theorem extends to higher dimensional spaces, and when it is a close approximation it provides the flexibility of inducing equivalent group actions no matter where the closed loop starts or stops.
Furthermore, any path in the kernel distribution of $A(\cdot)$ can connect such loops to one-another without introducing an additional motion in the group.
In practice, however, obtaining this kernel from data requires cumbersome sampling and system identification.

\subsection{Representational simplifications}

Given a gait $\gamma_b$, the body frame motion $M_b$ it produces could, in principle, be a function of the initial point in the gait cycle and the speed with which this cycle is executed.
For systems where momentum is dominated by friction or constraints (Stokesian systems), this is not the case.
In those systems there exists a map $A(r):r \in \basespace \mapsto \mathsf{L(\mathsf{T}_r,\aG)}$ taking shapes to linear maps from shape velocities to the Lie Algebra $\aG$ of $\fiberspace$.
This leads to the ``reconstruction equation'' $\dot g = L_g A(r) \dot r$ where $L_g: \aG = \mathsf{T}_e \to \mathsf{T}_g$ is the lifted left action of the group element $g$ (commonly written $g^{-1}\dot g = A(r) \dot r$ for matrix Lie groups).
Thus, if two base loops are connected at any point, the combination of their actions can be represented as a group multiplication of their respective $g$ elements.

There are infinite ways to take a gait library and coordinate it into a complete motion planner.
Typically people have a scheme for transitioning between gaits.
The overhead of finding such transitions for systems with no model can be large.

When selecting a collection of gaits for computing coverage, we required that each shares a common point in the base space and thereby allowing gait cycles to be applied in any order.

\section{Setup for Discovering a High Coverage Gait Library}

\label{sec:cov-results}
To illustrate our approach on a classical system, we simultaneously optimized three gaits on Purcell swimmers to provide coverage of a portion of $\mathsf{SE}(2)$ surrounding the identity using their $h_4(\cdot)$ cost.
\subsection{Coverage point selection}
\label{sec:unif}
This coverage point distribution included equally weighted points derived from all possible combinations of the following values, totaling 125 points:
\begin{align}
    x &= [-1,-0.5,0,0.5,1] \\
    y &= [-1,-0.5,0,0.5,1] \\
    \theta &= [-\pi,-\frac{\pi}{2},0,\frac{\pi}{2},\pi]
\end{align}
where units for translation were body lengths and units for rotation were in radians.
These spanned the translational bounds of moving by one body length and the rotation bounds of rotating by a half of a full rotation.

\subsection{Model extraction and motion parametrization} \label{sec:cov-purcell-setup}
A single iteration of learning involved experimentally running each of the three gaits for 30 noisy cycles, modeling their dynamics via the framework of \cite{bittner2018geometrically}.
We parametrized the gaits with a modified version of the ellipse with bump function parametrization also used in \cite{bittner2018geometrically}.
The following parametrization $p$ is a modification that allows the base point, $b_i$, of the three gaits to be an explicit parameter:

\begin{align}
r_i(t) := &c_{i} + (b_{i}-c_{i}) \cos(\Omega t) + a_{i} \sin(\Omega t) + \\ &\sum_{k=0}^{N_o-2} u_{i,k} \, w\left( t - k \frac{2 \pi}{N_o}\right)\label{eq:wind} \nonumber \\
w(x) := &\begin{cases}
  1+\cos(x f N_o) & |x f N_o| < \pi \\
  0 & |x f N_o| \geq \pi
    \end{cases},
\end{align}
with gait parameters
\begin{equation}
p_{i} = (c_{i}, b_{i}, a_{i}, u_{i,k}).
\end{equation}
In this work, we used $N_o$ = 18 and $f$ = 3, totaling 16 bumps.
Two bumps were elided ($k=17,18$) via this representation such that base point $b$ is left unshifted.

\section{Finding Coverage with Purcell swimmers}
\label{sec:covopt}

Our first investigation was to see how well Purcell swimmers can optimize three gaits simultaneously for the uniformly distributed set of coverage points of \S\ref{sec:unif}.
We observed how the ability to optimize these gaits changed as we added joints to the swimmer.
We started with two joints (the two-joint Purcell swimmer) and built our way up to eight joints.
We repeated the optimization process 30 times for each swimmer.

\begin{figure*}
\begin{center}
\includegraphics[width=\textwidth]{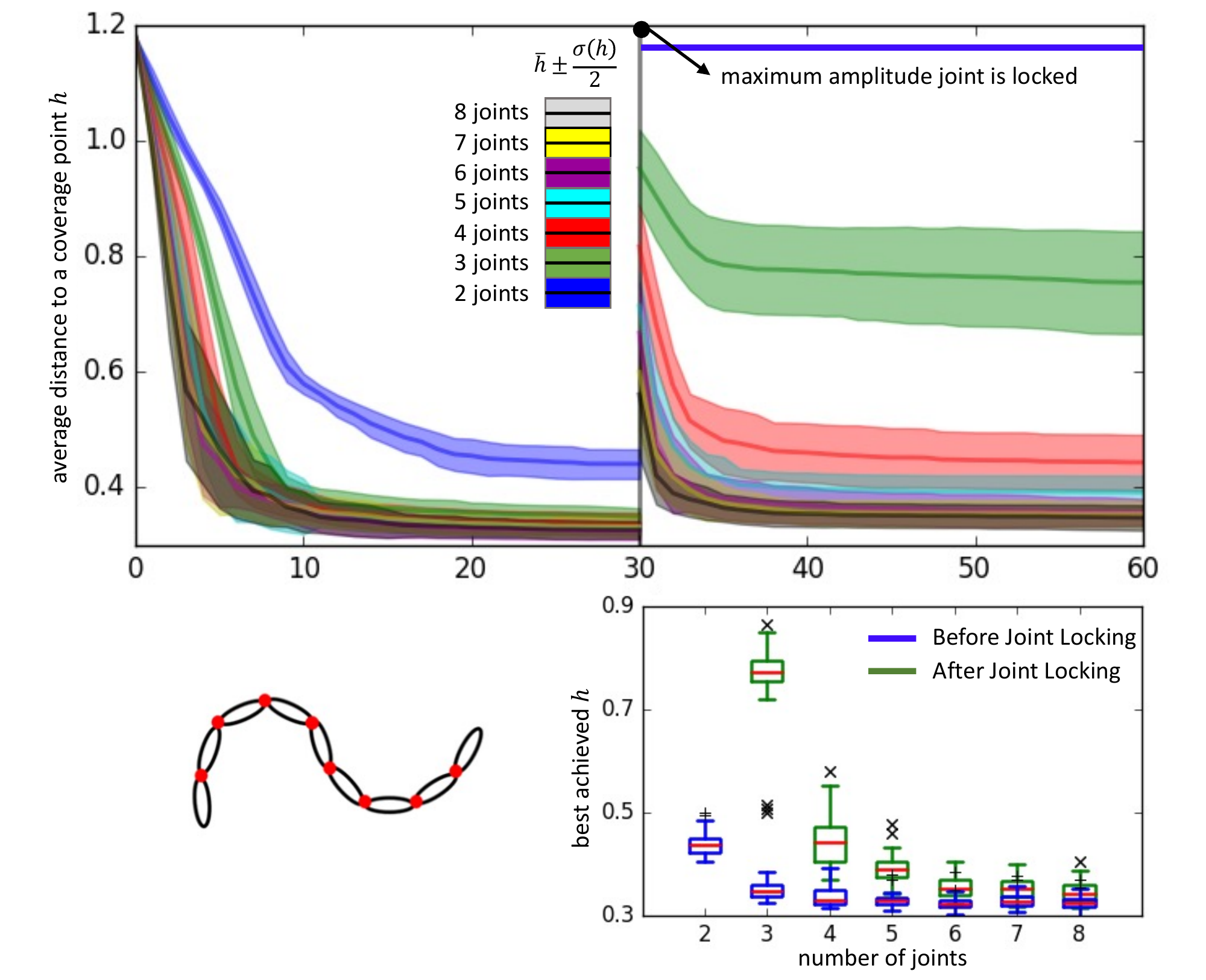}
\caption{Purcell swimmers of varying complexity, such as the eight-joint (pictured bottom left) were optimized for three gaits that maximize coverage. See \S\ref{sec:unif} and \S\ref{sec:cov-purcell-setup} for details on the setup of the experiment.
We plotted the mean (top, solid lines) and standard deviation (transparent bands) over 30 separate simulations of the average distance of goal motions to the nearest available motion, denoted $h$.
We can see how $h$ changes across trials and the number of joints used by the swimmer (2=blue, 3=green, 4=red, 5=cyan, 6=magenta, 7=yellow, 8=black).
At iteration 30 (marked by a vertical grey line), we plotted how well the swimmers adapt to having the maximal amplitude joint locked.
We also observed how the quality of the coverage of the library varies by the number of joints used by the swimmer (bottom right) before (blue box plots) and after (green box plots) joint locking.
}
\label{fig:learning}
\end{center}
\end{figure*}

At the beginning of each optimization, a random joint was stimulated with a sine wave.
The stimulated joint was distinct for each gait.
The only exception to this was that for the two-joint Purcell swimmer, a gait had to be repeated since there were three initial gaits and only two joints.
The swimmers used 30 cycles at each gait to build a model.
Then, the swimmers used the models provided by \cite{bittner2018geometrically} to predict how changing the parametrization of their three gaits could be combined to optimize a 4-step plan over the coverage points provided\footnote{Using 4 steps allowed us to include knowledge of the commutator motions noted in \S\ref{sec:commutator}.}.
An iteration of the optimization involved stepping along the policy gradient of three gaits (step size computed via \cite{bittner2018geometrically}) and simultaneously updating the three gaits. The results are recorded in Figure \ref{fig:learning}.

We ended the optimization after 30 iterations.
The test showed that the swimmers were able to use the coverage metric to consistently find a gait library for local motion planning.
Having two joints was sufficient for finding a gait library, but having three joints presented a notable improvement in coverage.
This jump in performance was less surprising after considering that the third joint allowed the swimmer to become fully actuated (when in non-singular configurations) with respect to $\mathsf{SE}(2)$.
After the third joint was added, the convergent behavior of the swimmers was consistently within the performance noise window of adding another joint, i.e. the marginal benefit of adding a joint was small.

The convergence rate of the swimmers improved when adding the third and fourth joints.
For all swimmers containing 3 or more joints, the standard deviation of performance reached $h=0.4$ by the tenth trial.
Here, we calculated $h$ as the average normed distance to a coverage point.
Converging at this expedient rate required exactly 900 cycles of robot data.
If we ran physical robots at 3Hz, the optimizations would have converged after collecting just five minutes of experimental data, even on the eight-joint swimmers.

\begin{figure*}
\begin{center}
\includegraphics[width=\textwidth]{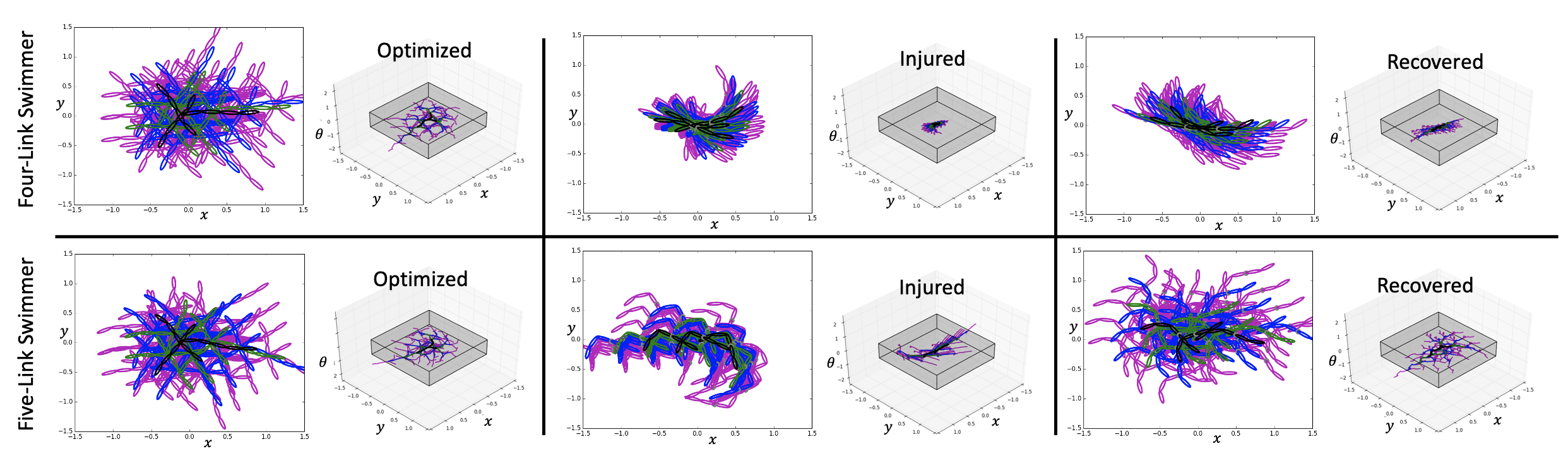}
\caption{This provides a detailed look at two optimization process for a three-joint and four-joint swimmer in the study summarized in Figure \ref{fig:learning}.
We plotted the 4 step horizon (1=black, 2=green, 3=blue, 4=magenta) at various trials on the plane (left in each section) and on $\mathsf{SE}(2)$ (right in each section).
For reference, we plotted the unit volume in $\mathsf{SE}(2)$ (gray box) over which the coverage points were uniformly distributed.
For the three-joint swimmer, we showed the optimal policy before injury in trial 22 (top right), the consequence of a locked joint (grey dot) on the optimal policy in trial 30 (top middle), and the optimal policy recovered while the joint remains locked in trial 52 (bottom right).
The three-joint swimmer was strongly impeded in its ability to recover a high coverage collection of gaits post-injury.
For the four-joint swimmer, we showed the optimal policy before injury in trial 17 (top right), the consequence of a locked joint (grey dot) on the optimal policy in trial 30 (top middle), and the optimal policy recovered while the joint remains locked in trial 54 (bottom right).
The four-joint swimmer was not impeded in its ability to recover a high coverage collection of gaits post-injury.}
\label{fig:recovery}
\end{center}
\end{figure*}

\section{Investigating the Ability of the Purcell Swimmer to Recover from Joint Locking}
\label{sec:recovery}

In trials 30-60 of Figure \ref{fig:learning}, we tested the ability of the Purcell swimmer to recover from simulated damage.
We took the optimal collection of gaits from the first 30 trials and found the joint that used the highest amplitude behavior.
We locked this joint at its value taken at the base point of the parametrization.
We then used this damaged optimal policy from the first 30 trials to start a on optimization leaving the damaged joints locked.

The two-joint swimmer was unable to move as a result of the injury.
The three-joint swimmer was able to partially recover.
It was equipped with two functional joints, yet was did not achieve the coverage scores of the un-injured two-joint Purcell swimmer.
The four-joint swimmers were notably better at finding high coverage libraries during recovery than the three-joint swimmers and remained within the standard deviation of performance of the five-joint swimmers.
The top row of Figure \ref{fig:recovery} details one optimization process for a swimmer with three joints.
It is clear that before injury, the swimmer was able to achieve local poses.
The injury greatly handicapped this ability, even with the opportunity to recover.
Likewise, the bottom row of Figure \ref{fig:recovery} details one optimization process for the swimmer with four joints.
Before injury, the four joint swimmer also found a useful gait library.
The injury clearly hindered its ability to move, but given the opportunity to recover, the four joint swimmer found a new collection of behaviors with good coverage.

We interpret these results as follows.
After damage, the two-joint swimmer was left with one joint and was unable to move, possibly a consequence of the scallop theorem in Stokesian systems \cite{lauga2011life}.
The three-joint swimmer recovered poorly, and did not come to match the performance of the undamaged two-joint swimmer.
This may suggest that the injury resulted in a body geometry that was less amendable to producing good coverage than the conventional two-joint Purcell swimmer.
As we added more joints, the redundancy of joints both minimized the dynamical impact of injury and provided a larger space of solutions for recovery.
The boxplots in Figure \ref{fig:learning} suggest that at around five or six joints, the coverage performance of the swimmer becomes robust to the locking of a single joint.

Since all recovery processes took the same amount of time, we have a rare result: adding actuated degrees of freedom improved our convergence and recovery rates.
Adding more freedom typically involves a substantial increase in sampling requirements, both lengthening the convergence process and making it less certain.
In this example, convergence rate either improved or stayed approximately the same as joints were added.
Here, combining the methods of \cite{bittner2018geometrically} and the coverage metric allowed redundancy in the internal state to be an asset for behavior optimization rather than a liability.

\section{Implementation on Hardware}

Here we communicate the general and noise-robust qualities of our approach by optimizing coverage on real hardware with an unknown model.
We did not have explicit knowledge of the kinematics, mass distribution, or material properties of this system when running the modeling and optimization algorithms.

\begin{figure*}
\begin{center}
\includegraphics[width=\textwidth]{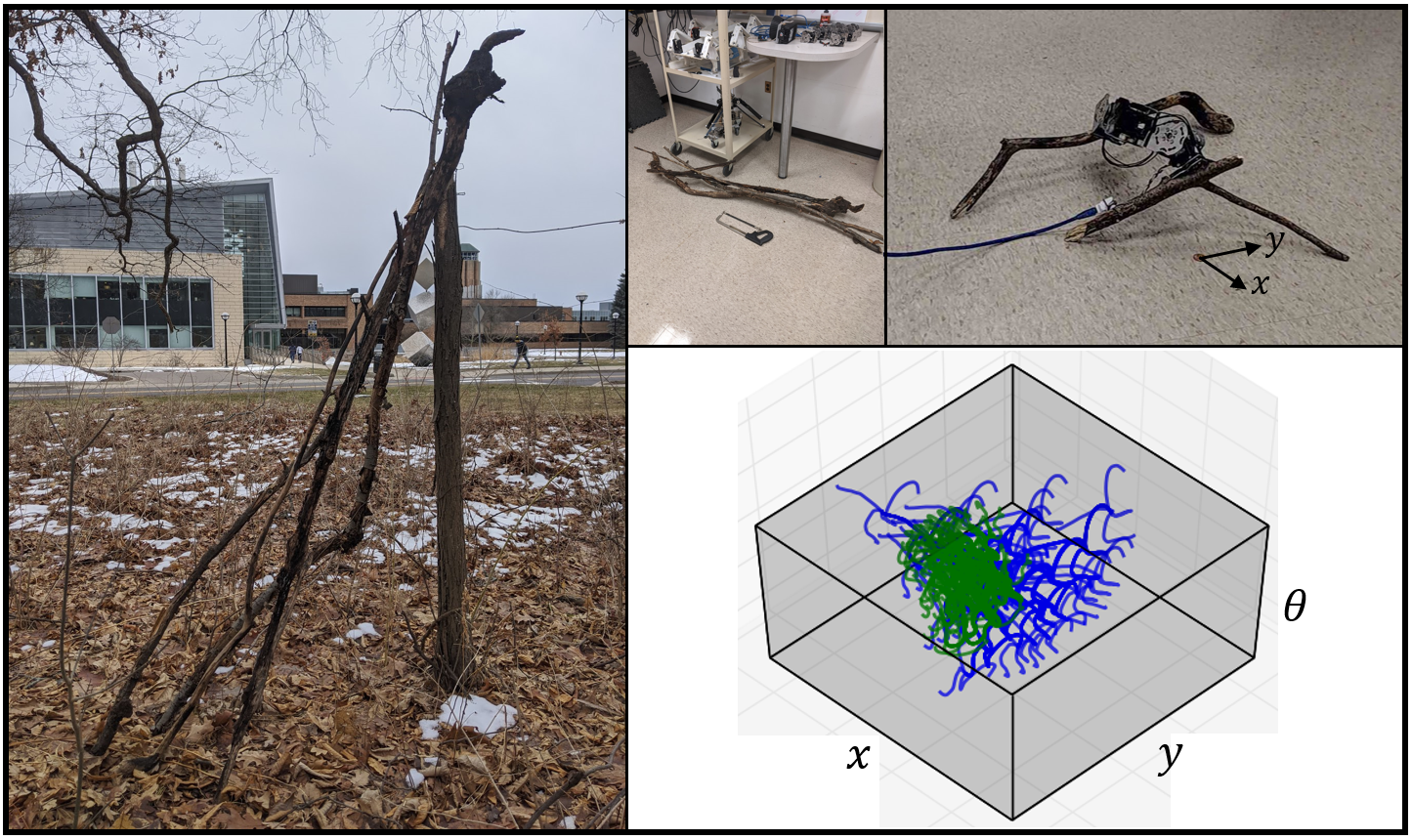}
\caption{This robot (top right) was built from dynamixel modules and tree branches available nearby (left and middle left).
The trajectories showcase the available 1 to 4 cycle motions of the system (bottom right) from the robot's origin before (green) and after (blue) the coverage optimization.}
\label{fig:creaturebot}
\end{center}
\end{figure*}

\subsection{Methods on hardware}
Inspired by the hardware used in \cite{maekawaimprovised}, we foraged for tree branches.
We gathered these and sectioned the branches into robot appendages of a useable size.
We then constructed a robot by fixing tree branches to the endpoints of a chain of three Robotis Dynamixel RX-64 actuators in modular cages.
We equipped the robot with 3 markers for our motion capture system (Qualisys Oqus 6 camera system); this provided us a observation of the position and orientation of the robot.
We connected the robot to a computer (Intel Xeon CPU E3-1246 v3 running at 3.50GHz) running the gait modeling and optimization algorithms.
This connection used a CAT5 cable with 3 conductor pairs for power and one pair for RS-485 serial communications.
To build a physics model centered at a given gait, we collect 20 cycles of noisy input data on the robot and fit a regression informed by physics and geometry \cite{bittner2018geometrically}.
We then compared the outcomes of two different optimizations for the $(x,y,\theta)$ outcomes of a gait or gaits, taking the position of the robot prior to application of a gait cycle as the origin.

We performed two gait optimization experiments:

\noindent \textbf{(1) Find one gait to move forward without turning:} We designed a gait optimization to maximize $x-y^2-\theta^2$ (per cycle) given the coordinates of Figure \ref{fig:creaturebot} and units of body lengths ($\frac{1}{3}m$) and radians.

\noindent \textbf{(2) Find three gaits that optimize coverage in a volume of $\mathsf{SE}(2)$}: Given the ability to use the 3 gaits in up to 4 combined cycles, we designed a gait library optimization to minimize the distance (computed on the Lie group), from 125 points distributed across all combinations of coordinates $x=[-1,-\frac{1}{2},0,\frac{1}{2},1]$, $y=[-1,-\frac{1}{2},0,\frac{1}{2},1]$, $\theta = [-\frac{\pi}{2},-\frac{\pi}{4},0,\frac{\pi}{4},\frac{\pi}{2}]$.
The gray volume in Figure \ref{fig:creaturebot} contains all of the coverage points.

\subsection{Results on hardware}
For the first goal function, we seeded a zero motion gait oscillating the middle joint with a sinusoidal input.
We executed 15 iterations of our data-driven gait optimization algorithm, each consisting of 20 cycles of motion.
Running at $\frac{1}{2}$Hz, each trial took 40 seconds.
After the 8th iteration, the robot was able to travel $40\%$ of its body length per cycle with a turning rate of $0.10$ radians per cycle.

For coverage, we first completed an exploratory sampling of motions (12 cycles).
From these 12 different gaits, we selected the subset of 3, which performed the best on the coverage metric.
After 5 iterations of trials (60 cycles, 20 for each gait), the system found a more complementary set of gaits reducing the coverage score from 0.97 to 0.76.

\section{Discussion and Conclusions}

In this paper, we introduced a new metric for the optimization of robot motions.
This metric involved calculation of the composition of motions from a small library of primitives, determining their utility in ``covering'' some region of the local body position space, formulated as a Lie group.
What is novel about this approach is that
\begin{itemize}
\item It eliminated human bias from prescribing a limited set of allowable primitives for a robot.
\item It allowed for the use of unconventional robot designs for navigation.
\item It allowed malfunctioning robots to quickly recover the ability to move through space.
\end{itemize}

We showed the Purcell swimmers' ability to recover from injury using the data-driven geometric gait optimizer, guided by the coverage metric.
Some interesting trends emerged during these tests.
The swimmers converged to a high coverage gait library (containing three gaits) despite variation in the number of links and initial gaits.
This suggests insensitivity in the gait optimization when using the coverage metric.

Furthermore, coverage allowed us to investigate the role that redundancy might play in the ability of the swimmers to recover high coverage gait libraries post-injury.
We found that around four degrees of freedom, the addition of a joint no longer provides a substantial change in the ability of the swimmer to recover.
The ability to apply this analysis to other robots could help inform what degree of complexity is appropriate when designing a robot, and provide a lower bound for how much recovery can be expected from different amounts of damage.

Finally, using the coverage optimization on a robot made of tree branches we were unable to find gaits that translate without substantial rotation.
We were able, however, to find a useful portfolio of maneuvers for navigation in 2D.
This machine learning task was solved on a timescale that is competitive with an implementation of reinforcement learning by Google \cite{ha2020learning}.

The tree branch robot example speaks to the morphology agnostic properties of data-driven geometric gait optimization; this robot could be substituted with robots of many other forms.
As long as the system acts near the Stokesian regime of locomotion, the methods of \cite{bittner2018geometrically} assist in building behavioral models that inform performance improvements.
Interfacing the coverage optimization metric to soft systems (where approaches to system identification have been developed \cite{bruder2020data,bittner2020data}) could enable more reliable soft robots in hard to model environments.

Taken together our results give strong evidence that optimizing for coverage is a means for a robot to gain the ability to maneuver, and to recover this ability after being damaged.

\section*{Acknowledgement}
Revzen and Bittner thank NSF CMMI 1825918, ARO W911NF-17-1-0306, D. Dan and Betty Kahn Michigan-Israel Partnership for Research and Education Autonomous Systems Mega-Project, and AOFSR FA9550-20-1-0238 for funding.

\printbibliography
\end{document}